\documentclass[11pt]{article}
\usepackage{last-preprint}
\usepackage{graphicx}
\usepackage{booktabs,multirow,array}
\usepackage{enumitem}
\usepackage{pifont}
\usepackage{float}

\title{LaST$_{1.0}$: A Unified Latent World-Action Model\\
with Control-Centric SpatioTemporal Reasoning}
\shorttitle{LaST$_{1.0}$: A Unified Latent World-Action Model with Control-Centric SpatioTemporal Reasoning}
\author{%
  Hao Chen\textsuperscript{1,*,$\dagger$}\quad
  Xuheng Zhang\textsuperscript{2,*}\quad
  Zhuangzhe Wu\textsuperscript{2,*}\quad
  Jiaming Liu\textsuperscript{2}\quad
  Shaobo Zhang\textsuperscript{2}\quad
  Jialin Gao\textsuperscript{1}\quad
  Zhonghao Yan\textsuperscript{1}\\[4pt]
  Di Wu\textsuperscript{2}\quad
  Sixiang Chen\textsuperscript{2}\quad
  Peng Jia\textsuperscript{3}\quad
  Chi-Wing Fu\textsuperscript{1}\quad
  Shanghang Zhang\textsuperscript{2,\ding{41}}\quad
  Pheng-Ann Heng\textsuperscript{1}\\[9pt]
  {\small\textsuperscript{1}The Chinese University of Hong Kong\qquad
   \textsuperscript{2}State Key Laboratory of Multimedia Information Processing,}\\[2pt]
  {\small School of Computer Science, Peking University\qquad
   \textsuperscript{3}Simplexity Robotics}\\[7pt]
  {\footnotesize\textsuperscript{*}Equal contribution\quad
   \textsuperscript{$\dagger$}Project leader\quad
   \textsuperscript{\ding{41}}Corresponding author}%
}
\date{}
\hypersetup{pdftitle={LaST1.0: A Unified Latent World-Action Model with Control-Centric SpatioTemporal Reasoning},
  pdfauthor={Hao Chen, Xuheng Zhang, Zhuangzhe Wu, Jiaming Liu, Shaobo Zhang, Jialin Gao, Zhonghao Yan, Di Wu, Sixiang Chen, Peng Jia, Chi-Wing Fu, Shanghang Zhang, Pheng-Ann Heng}}

\begin{document}
\maketitle

\begin{abstract}

World-Action Models couple action learning with future prediction, enabling robotic policies to capture environmental dynamics for robust control. However, prevailing methods take a reconstruction-oriented approach to model future videos, allocating substantial capacity to appearance details that are incidental to control. In this work, we reformulate world modeling for manipulation as control-centric future prediction in latent space, and present \textbf{LaST$_{1.0}$}, a unified latent World-Action Model. Rather than modeling an observation-complete video future, LaST$_{1.0}$ predicts compact, geometry-aware latent targets derived from future wrist views, focusing the supervision on the evolving robot--object interactions required for fine-grained manipulation. To better support this control-centric paradigm, LaST$_{1.0}$ revisits the visual input interface and attention mechanism of its unified architecture. It adopts a vision-encoder-free design with a lightweight patch embedder, exposing fine-grained visual cues directly to the backbone, and introduces LaST-Attention for structured spatiotemporal interactions across modalities. Evaluations across two robot embodiments, five challenging manipulation tasks, and three simulation benchmarks---RLBench, LIBERO, and LIBERO-Plus---demonstrate that LaST$_{1.0}$ achieves state-of-the-art success rates and generalizes to novel scenarios. These results demonstrate the effectiveness of control-centric latent world modeling for robust manipulation.

\end{abstract}

\begin{figure}[!htbp]
    \centering
    \includegraphics[width=\textwidth]{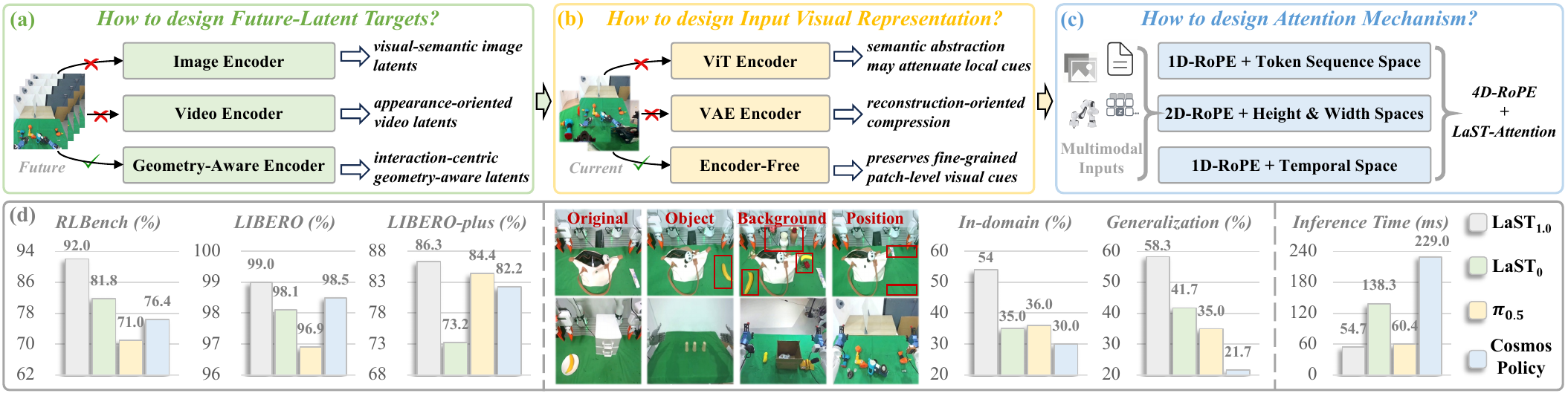}
    \caption{\textbf{Overview of LaST$_{1.0}$.} The top panel illustrates the design roadmap of LaST$_{1.0}$, highlighting the interaction-centric, geometry-aware future-latent supervision, alongside early fusion of visual patches and structured 4D spatiotemporal attention. The bottom panel summarizes the model's performance across simulation, real-world, generalization, and inference latency.}
    \label{fig:teaser}
\end{figure}

\section{Introduction}

Vision-Language-Action (VLA) models have become a strong foundation for generalist robot manipulation by directly mapping visual observations and language instructions to actions \cite{intelligence2025pi05visionlanguageactionmodelopenworld,liu2025hybridvla,kim2025fine,liu2026last}. Despite their strong visual and semantic capabilities, the imbalance between dense visual inputs and sparse action supervision can lead to shortcut mappings rather than learning underlying physical dynamics \cite{ye2026gigaworld,luo2026being}.
Consequently, these models can struggle with robust control in complex scenarios that require anticipation of future interactions.
World-Action Models (WAMs) address this limitation by coupling action learning with future-state prediction, using future prediction as an auxiliary objective that encourages the shared representation to capture task-relevant changes in the environment \cite{ye2026world,cheang2024gr,kim2026cosmos,team2026gigaworld}.

A central question, however, is \emph{what} a policy model should predict about the future. Many existing WAMs instantiate future-state modeling through video generation \cite{ye2026gigaworld,ye2026world,yuan2026fast,kim2026cosmos}.
While visually expressive, this paradigm allocates substantial capacity to texture, illumination, background appearance, and other photometric details that have little bearing on the next physical interaction \cite{luo2026being}.
In contrast, fine-grained control depends more directly on localized object motion, relative geometry, contact configuration, and how the interaction region between robot-object evolves over time.

We therefore propose that world modeling for manipulation should be \emph{control-centric} rather than \emph{observation-complete}. For manipulation, the future latent representation should satisfy three properties: (1) preserve the geometric evidence needed for precise action, (2) focus modeling capacity on the region where interaction occurs, and (3) provide predictive supervision during training without becoming mandatory computation at deployment. This perspective changes the role of future prediction: the goal is not to reproduce the world in all observable detail, but to predict the subset of future dynamics that is useful for control.
Guided by this principle, we introduce \textbf{LaST$_{1.0}$}, a unified latent WAM that jointly optimizes action generation and control-centric future dynamics prediction. Rather than predicting appearance-oriented video latents or visual-semantic image latents, LaST$_{1.0}$ utilizes a geometry-aware encoder \cite{wang2026vggt} to construct compact targets exclusively from future wrist-camera observations, as described in Fig.~\ref{fig:teaser}(a). This concentrates the latent supervision on the interaction region, forcing the policy to capture fine-grained relative motions and robot--object geometric relationships. Crucially, our experiments reveal that formulating latent world modeling as a training-time regularizer not only yields improved performance but also enhances closed-loop control efficiency.

To operationalize this control-centric paradigm, we further redesign the visual interface and attention mechanism for LaST$_{1.0}$, as shown in Fig.~\ref{fig:teaser}(b) and (c). Semantic vision encoders and reconstruction-oriented VAEs can bias visual representations toward high-level semantics or appearance reconstruction, potentially attenuating the localized visual details vital for manipulation \cite{luo2026look,wang2026repwam,liu2026tuna,diao2026pixels,wei2026pixelumm}. LaST$_{1.0}$ therefore adopts a vision-encoder-free interface, using a lightweight patch embedder to project raw image patches directly into the unified Transformer backbone, thereby preserving fine-grained visual cues. Within this backbone, rather than relying on conventional attention mechanisms, we introduce LaST-Attention to facilitate the spatiotemporal fusion of multimodal information across 4D subspaces (token sequence, image height, image width, and relative time), effectively supporting our latent world-action modeling paradigm.

For evaluation, we deploy LaST$_{1.0}$ across two distinct embodiments, utilizing both parallel grippers and dexterous hands across five challenging manipulation tasks. These tasks evaluate the model's ability to perform fine-grained control, spatial reasoning, and long-horizon manipulation. Empirical results demonstrate that LaST$_{1.0}$ achieves state-of-the-art (SOTA) success rates, exhibiting remarkable generalization to novel object, background, and position settings. Moreover, evaluations on three benchmarks---RLBench, LIBERO, and LIBERO-Plus---confirm that LaST$_{1.0}$ outperforms prior SOTA baselines, further validating the efficacy of our latent world-action modeling paradigm, as shown in Fig.~\ref{fig:teaser}(d).
The main contributions of this work are as follows:
\begin{itemize}[nosep, leftmargin=*]
\item  We reformulate world modeling as the prediction of geometry-aware, control-centric future dynamics in latent space, using latent targets derived from future wrist views to capture evolving robot--object interactions.
\item  We introduce LaST$_{1.0}$, a latent World-Action Model that natively unifies vision-language understanding, latent world modeling, and action generation. Within this model, we design a vision-encoder-free interface and LaST-Attention to preserve fine-grained spatial cues and facilitate structured spatiotemporal fusion across modalities.
\item  Experiments demonstrate that LaST$_{1.0}$ achieves state-of-the-art performance across two embodiments, five challenging manipulation tasks, and three simulation benchmarks, together with robust generalization to novel scenarios.
\end{itemize}

\section{Related Work}

\subsection{Vision-Language-Action and World-Action Models}
Generalist robot policies have primarily evolved along two complementary paradigms. Vision-language-action (VLA) models directly map visual observations and language instructions to robot actions \cite{intelligence2025pi05visionlanguageactionmodelopenworld,liu2025hybridvla,kim2025fine,chen2025fast,black2024pi_0,kim2024openvla,ye2026starvla,gu2025manualvla,liu2026last,wu2026pragmatic,nvidia2025gr00tn1openfoundation,liurdt,team2026xiaomi,li2026sla,fang2026molmoact2,li2026spatial,qu2025spatialvla,cen2025worldvla}. World-action models (WAMs) additionally model future visual dynamics, either jointly with action prediction or through an imagine-then-act formulation, to provide predictive priors for control \cite{cheang2024gr,li2025unified,bi2026motus,yuan2026fast,nvidia2026gr00t2,li2026causal,ye2026world,huang2026long,guo2026video,kim2026cosmos}.
To focus on more informative representations and improve inference efficiency, recent works have explored world modeling in latent spaces \cite{luo2026being,sun2026vla,chen2026last,lyu2026lda,wang2026st}. However, these approaches often emphasize scene-level representations rather than information directly relevant to robot-object interaction and manipulation control. LaST$_{1.0}$ instead focuses latent world modeling on control-centric information by constructing geometry-aware latent targets that capture the evolving robot-object interaction, providing more relevant predictive supervision for action generation. To better support this control-centric latent reasoning, we further co-design the visual interface and the unified backbone. An encoder-free patch interface preserves fine-grained interaction cues, while LaST-Attention with 4D-RoPE enables multimodal fusion that respects spatial and temporal structure. Task-routed experts specialize in context understanding, latent dynamics modeling, and action generation, providing complementary capacity for learning control-relevant predictive representations.

\subsection{Geometric Representation for Robotic Manipulation}
Geometric information is essential for precise manipulation, where actions depend on spatial relations and robot--object configurations. Existing approaches incorporate geometry through explicit inputs, feature fusion, or representation supervision. DepthVLA \cite{yuan2025depthvla} enriches policies with depth information, while point-cloud-based VLAs, including PointVLA \cite{li2026pointvla} and GeoVLA \cite{sun2025geovla}, integrate 3D features into action prediction \cite{liu2025mla,li20253ds}. BridgeVLA \cite{li2025bridgevla} instead projects 3D observations into multiple 2D views, aligning visual inputs and action heatmaps within a shared image space. FALCON \cite{zhang2025falcon} supplies an action head with spatial tokens from geometric foundation models, while Evo-0 \cite{lin2025evo} incorporates geometry-aware features extracted from RGB observations. SpatialVLA \cite{qu2025spatialvla} further introduces egocentric 3D position encodings during policy pretraining. These designs retain depth estimation, point-cloud processing, or geometric feature extraction in the inference pipeline.
A complementary direction internalizes spatial structure through training-time supervision. Spatial Forcing \cite{li2026spatial} aligns intermediate policy features with geometric representations, and ROCKET \cite{sun2026rocket} extends alignment across multiple backbone layers through a shared projector. VEGA \cite{wang2026vega} directly supervises visual encoder outputs, while Mind-VLA \cite{ding2026mindvla} uses instruction-aware target-object representations to focus geometric supervision on the manipulated object. These approaches strengthen spatial perception without requiring the auxiliary geometric modules at deployment. Beyond spatial representation alignment, 4D-WAM \cite{yang20264d} transfers spatiotemporal knowledge from 3D trajectory fields through motion and destination alignment. LaST$_0$ \cite{liu2026last} also uses future point-cloud features as predictive supervision. LaST$_{1.0}$ advances this direction by predicting compact geometry-aware latents extracted from future wrist-camera observations, concentrating supervision on evolving local robot--object interactions without requiring explicit point-cloud data. Its branch-decoupled training makes future geometry a predictive regularizer, allowing both the geometric teacher and the latent prediction branch to be omitted at inference.

\begin{figure}[tbp]
    \centering
    \includegraphics[width=\textwidth]{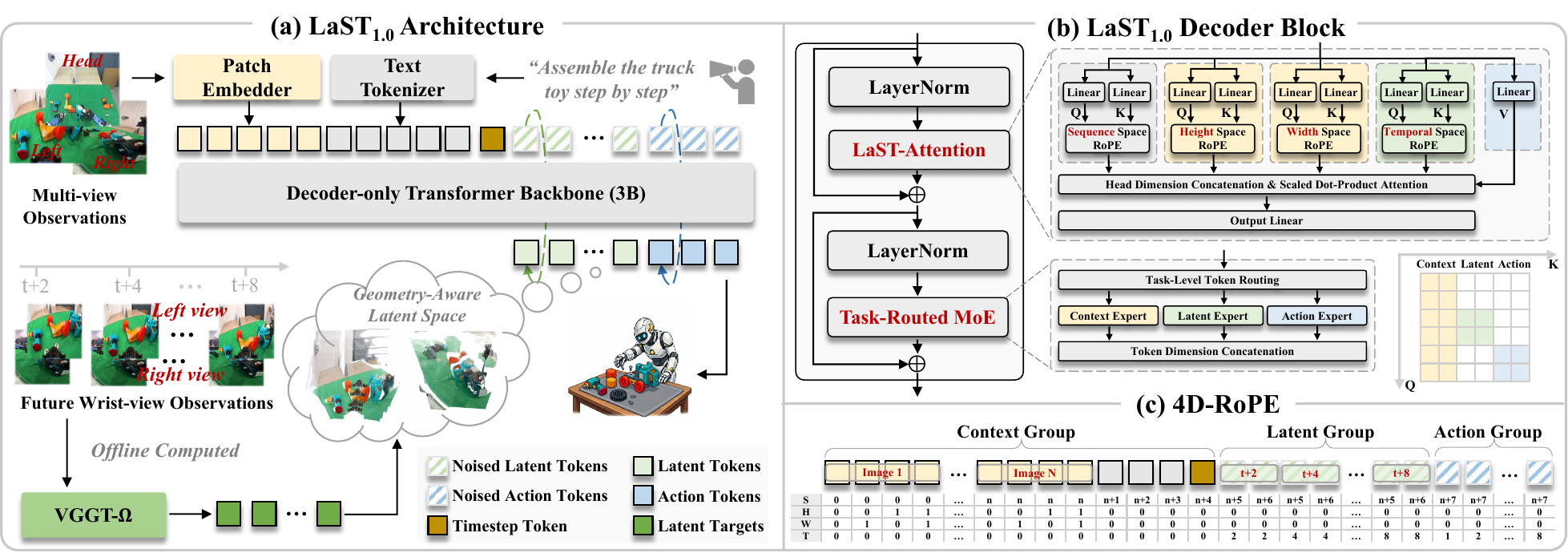}
    \caption{\textbf{LaST$_{1.0}$ Framework.} \textbf{(a)} LaST$_{1.0}$ employs an encoder-free unified backbone that takes projected multimodal context to jointly predict flow velocities for latent and action tokens. The latent targets are extracted via VGGT-$\Omega$ from future wrist views to enable interaction-centric geometry-aware reasoning. \textbf{(b)} Each decoder block in LaST$_{1.0}$ features LaST-Attention, which independently rotates queries and keys across token sequence, image height, image width, and time axes to enhance spatiotemporal fusion. A subsequent Task-Routed MoE effectively expands model capacity. \textbf{(c)} Multimodal Tokens are represented in a 4D space with separate sequence, spatial and temporal coordinates, preserving their inherent structure.}
    \label{fig:method}
\end{figure}

\section{Method}
\label{sec:method}

We first formulate the problem in Sec.~\ref{sec:problem_formulation}, then introduce the architecture of LaST$_{1.0}$ and LaST-Attention in Secs.~\ref{sec:last_architecture} and \ref{sec:last_attention}, respectively. We describe control-centric latent world modeling in Sec.~\ref{sec:latent_world_modeling} and the training and inference procedures in Sec.~\ref{sec:training_inference}.

\subsection{Problem Formulation}
\label{sec:problem_formulation}
We formulate vision-language robotic manipulation as learning a policy $\pi_\theta$ via imitation learning on expert demonstrations $D$.
At each timestep $t$, given the language instruction $l$ and multi-view observations $I_t=\{I_t^k\}_{k=1}^{K}$, the policy predicts an action chunk $\mathbf{a}_{t+1:t+H}$ over a horizon $H$. Each action encodes an embodiment-specific control command, comprising a relative end-effector pose coupled with either a parallel gripper state or a dexterous multi-finger configuration.
To regularize action learning, we formulate future observations as an auxiliary latent world-modeling target (detailed in Sec.~\ref{sec:latent_world_modeling}). This predictive supervision acts as a training-time regularizer, embedding physical priors into the representation while allowing the policy $\pi_\theta$ to execute closed-loop control based solely on $(l, I_t)$ at inference.

\subsection{\texorpdfstring{LaST$_{1.0}$}{LaST1.0} Architecture}
\label{sec:last_architecture}
LaST$_{1.0}$ features an encoder-free unified architecture, projecting multimodal signals directly into the backbone to preserve fine-grained visual details critical for manipulation. To facilitate effective multimodal fusion without overwhelming the backbone's capacity, we employ task-routed experts ensuring stable and decoupled learning across vision-language understanding, latent world modeling, and action generation.

\subsubsection{Encoder-Free Multimodal Projection}
The language instruction $l$ is first encoded by the pretrained Qwen3-1.7B tokenizer \cite{yang2025qwen3} and mapped into embeddings $f^{\text{lang}} \in \mathbb{R}^{N_{\text{lang}} \times d_h}$ with a hidden dimension of $d_h=2048$.
For visual inputs, while conventional deep vision encoders typically yield tokens that focus on high-level semantic information \cite{luo2026look,kachaev2025don,diao2026pixels,liu2026tuna,wei2026pixelumm}, precise robotic manipulation relies heavily on fine-grained local details. To better capture this information, LaST$_{1.0}$ bypasses deep vision encoders in favor of a lightweight projection interface, as shown in Fig.~\ref{fig:method} (a). Specifically, raw images $I \in \mathbb{R}^{H_{\mathrm{img}} \times W_{\mathrm{img}} \times 3}$, with height $H_{\mathrm{img}}$ and width $W_{\mathrm{img}}$, are processed by a two-layer convolutional module, which focuses on extracting local spatial details and directly projects them into the backbone's latent space, yielding visual tokens $f^{\text{img}} \in \mathbb{R}^{N_{\text{img}} \times d_h}$. These visual tokens are concatenated with the language tokens and a flow-matching timestep token to form the \textbf{context} group. Furthermore, the noised future-latent targets and action chunks are independently projected to form the \textbf{latent} and \textbf{action} groups. Logically, a single training sample is composed of these three distinct token groups:
$[\mathbf{X}^{\text{context}};\; \mathbf{X}^{\text{latent}};\; \mathbf{X}^{\text{action}}]$.

\subsubsection{Shared Backbone with Task-Routed MoE}
The backbone of LaST$_{1.0}$ is a 20-layer Transformer architecture, initialized with the first 20 layers of Qwen3-1.7B \cite{yang2025qwen3}. To efficiently scale model capacity, we replicate the standard Feed-Forward Network (FFN) in each layer threefold to construct a task-routed Mixture-of-Experts (MoE), as shown in Fig.~\ref{fig:method} (b). Specifically, the pretrained FFN is cloned into \emph{context}, \emph{latent}, and \emph{action} experts, deterministically routed to process their corresponding tokens. This design expands parameter capacity while activating only one FFN expert per token in each layer.

\subsubsection{Latent and Action Decoders}
Separate final layer normalizations and two-layer MLP heads decode the latent and action states into flow velocities:
$\widehat{\mathbf{V}}_A=G_A(\mathbf{H}^{A})$ and
$\widehat{\mathbf{V}}_Z=G_Z(\mathbf{H}^{Z})$.
Here, $G_A$ and $G_Z$ are branch-specific heads for action and future-latent tokens, respectively.

\subsection{LaST-Attention with 4D-RoPE}
\label{sec:last_attention}

While task-routed experts specialize the FFN computation, we retain shared attention to integrate information across modalities under the branch-decoupled mask. To make these interactions sensitive to spatial and temporal structure, we introduce LaST-Attention with 4D-Rotary Positional Embedding (RoPE). This design endows each token with four distinct positional dimensions and aggregates attention computations across these subspaces to support control-centric latent reasoning.

\subsubsection{4D-RoPE}
\label{sec:four_dimensional_coordinates}
As shown in Fig.~\ref{fig:method} (c), each token is assigned a 4D coordinate tuple $\mathbf{p} = (n, y, x, \tau)$ corresponding to its sequence order, spatial height, spatial width, and relative time. This coordinate allocation is strictly governed by token semantics:
Language tokens utilize only the sequence axis $(n, 0, 0, 0)$. Visual patches from the same image share a base sequence index and temporal offset $\tau$, but logically span the 2D grid $(y, x)$. Crucially, the generative latent and action tokens are assigned distinct combinations of sequence and temporal coordinates to uniquely identify their future slots and action horizons. This structured 4D allocation provides a stable coordinate convention across modalities and structurally prevents variable-length sequential inputs from distorting the underlying spatiotemporal relationships.

\subsubsection{LaST-Attention}
\label{sec:axis_specific_qk}
To effectively process the 4D coordinates assigned to each token, LaST-Attention partitions every attention head into four independent subspaces. Specifically, the queries and keys within a single head are projected into subspaces corresponding to the axis set $\phi=\{\mathrm{S}, \mathrm{H}, \mathrm{W}, \mathrm{T}\}$, which explicitly map to the sequence ($n$), height ($y$), width ($x$), and temporal ($\tau$) dimensions of the coordinate tuple $\mathbf{p}$. Axis-specific RoPE is then applied independently to each subspace based on its corresponding coordinate value.
Recognizing that different axes operate on vastly different scales, we assign distinct base frequencies ($\beta$) to each RoPE axis. The sequence axis inherits the large base ($\beta_{\mathrm{S}}=10^6$) from the LLM, while the spatial and temporal axes utilize progressively smaller bases ($\beta_{\mathrm{H,W}}=10^4, \beta_{\mathrm{T}}=10^2$) to reflect the substantially shorter coordinate ranges of image grids and action horizons.
After independent rotation, the four subspaces are concatenated along the head dimension to reconstruct the full query and key representations.
The final attention logit between token $i$ and $j$ is thus formulated as:
\begin{equation}
    e_{ij} = \frac{1}{\sqrt{d_{\mathrm{head}}}} \sum_{\alpha \in \phi} \left\langle \overline{\mathbf{q}}_{i}^{\alpha}, \overline{\mathbf{k}}_{j}^{\alpha} \right\rangle + M_{ij},
    \label{eq:last_attention_logit}
\end{equation}
where $\overline{\mathbf{q}}^{\alpha}$ and $\overline{\mathbf{k}}^{\alpha}$ denote the 4D-RoPE-injected subspace representations, and $\mathbf{M}$ is the attention mask (as shown in Fig.~\ref{fig:method} (b)). Specifically, $M_{ij}=0$ if token $i$ is allowed to attend to token $j$, and $-\infty$ otherwise.
These logits are subsequently passed through a softmax function to weight the value representations $\mathbf{v}$. Consequently, the final aggregated output of each attention head jointly reasons over relative sequence order, 2D spatial displacement, and temporal offset.

\subsection{Control-Centric Latent World Modeling}
\label{sec:latent_world_modeling}

While existing WAMs \cite{ye2026world,yuan2026fast} typically predict future video latents, our \emph{control-centric} formulation instead models a compact future interaction state. This approach retains information relevant to manipulation while discarding incidental appearance variation. Our latent targets exhibit three core properties: \textbf{(1) geometry-aware}, preserving the spatial structure that governs object motion and contact; \textbf{(2) interaction-centric}, focusing on the localized region where manipulation occurs; and \textbf{(3) inference-independent}, providing privileged future supervision during training without becoming a computational dependency at deployment.

\subsubsection{Geometry-Aware Future Targets}
\label{sec:geometry_targets}

Rather than conditioning the policy on explicit 3D point clouds, we construct a compact implicit geometric target using a frozen VGGT-$\Omega$ \cite{wang2026vggt}. Its register tokens provide a compact representation of the spatial structure within visual observations. By jointly encoding multiple future frames, the model produces temporally fused, geometry-aware latents instead of independent per-frame features. Crucially, rather than treating VGGT-$\Omega$ as a spatial feature extractor for \emph{current} observations, we repurpose it to construct geometry-aware \emph{future} targets.

\subsubsection{Interaction-Centric Wrist World}
\label{sec:wrist_world}
To reduce supervision from global background regions, we restrict future modeling to the \emph{wrist-world}, i.e., short-horizon future observations captured by the wrist cameras. This perspective centers the predictive target on the evolving geometric interactions between the end-effectors and manipulated objects.
By default, the action-chunk horizon is $H=8$, from which we uniformly sample $F=4$ future timesteps at offsets $\delta_f=\frac{fH}{F}=2f$ for $f=1,\ldots,F$, corresponding to $\{t+2,t+4,t+6,t+8\}$. For each arm, we collect its wrist-view frames across these $F$ sampled timesteps and input them together into the frozen VGGT-$\Omega$. The encoder extracts $R=16$ register tokens per frame, yielding a temporal latent representation of $16 \times F$ tokens for this specific arm, denoted as $\mathbf{Z}^{\text{arm}}$. Crucially, for dual-arm setups, we independently process this multi-frame wrist sequence for each arm, and then concatenate their respective latents along the token dimension to construct the complete future-latent target $\mathbf{Z} = [\mathbf{Z}^{\text{left}}, \mathbf{Z}^{\text{right}}]$.

\subsubsection{Training-Only Latent Regularization}
\label{sec:deployment_free}

Following \cite{yuan2026fast}, we use a branch-decoupled attention mask $\mathbf{M}$, as shown in Fig.~\ref{fig:method}(b), to formulate latent world modeling as a training-time regularizer. At every layer, context queries attend only to context keys, while latent and action queries each attend to the context and their own branch. This prevents action predictions from depending on future-latent tokens, either directly or indirectly through the context. The latent prediction loss nevertheless updates the shared context pathway and attention parameters, transferring geometric supervision to action learning through shared representations. This separation allows the latent branch to be omitted at inference, when actions are generated solely from the current context and noisy action tokens. Separately, we precompute the geometry-aware targets with the frozen VGGT-$\Omega$, avoiding repeated teacher inference during policy training.

\subsection{Training and Inference}
\label{sec:training_inference}

\subsubsection{Pre-training}
\label{sec:pre-train}

Our pre-training data preparation is adapted from LaST$_{0.5}$~\cite{zhang2026last0}, using only dual-arm parallel-gripper demonstrations. We construct the corpus from three data collections: AGIBot~\cite{bu2025agibot}, Simplexity Robotics i7, and GenRobot. As summarized in Table~\ref{tab:pretraining_data}, the combined corpus contains 255,760 trajectories, totaling approximately 3.9K hours and 139.5 million timesteps at 10\,Hz. Training windows are sampled uniformly over timesteps in the combined corpus, so each collection contributes in proportion to its number of timesteps.

\begin{table}[H]
    \centering
    \small
    \setlength{\tabcolsep}{6pt}
    \caption{\textbf{Pre-training data composition.}
    Durations are computed at 10\,Hz, with synchronized
    camera views counted as a single timestep.}
    \label{tab:pretraining_data}
    \begin{tabular}{lrr}
        \toprule
        Collection & Trajectories & Hours \\
        \midrule
        AGIBot                 & 153,717 & 2,534.8 \\
        GenRobot               & 36,849  & 1,026.9 \\
        Simplexity Robotics i7 & 65,194  & 313.2   \\
        \midrule
        Total                  & 255,760 & 3,874.9 \\
        \bottomrule
    \end{tabular}
\end{table}

Following LaST$_{0.5}$~\cite{zhang2026last0}, we align end-effector poses to a canonical right-handed convention, with the local $x$, $y$, and $z$ axes corresponding to the forward, gripper-opening, and upward directions, respectively. A per-source rotation reconciles native end-effector axes with this convention. Each trajectory uses a fixed world frame whose origin is the left arm's initial position; camera-frame poses are transformed into this frame using the corresponding extrinsics. End-effector poses are recovered through forward kinematics when they are not directly provided. The standardized state consists of the pose and scalar gripper opening for each arm.
We recompute actions from the aligned states instead of reusing source-provided action labels. Let $T_t\in\mathrm{SE}(3)$ denote an arm's standardized end-effector pose at timestep $t$, with rotation $R_t$ and position $\mathbf{p}_t$. The local relative action between consecutive resampled poses is
\begin{equation}
    \delta T_t=T_t^{-1}T_{t+1},\qquad
    \delta R_t=R_t^{\top}R_{t+1},\qquad
    \delta\mathbf{p}_t=R_t^{\top}(\mathbf{p}_{t+1}-\mathbf{p}_t).
    \label{eq:pretraining_relative_action}
\end{equation}
Both increments are thus expressed in the current end-effector frame. Source rotations, including quaternion annotations, are first converted to rotation matrices; the relative rotation $\delta R_t$ is then encoded by three Euler angles. For each gripper, an increase or decrease in opening beyond a threshold sets the command to open or close, respectively; otherwise, the previous command is retained. The first command is initialized from the measured opening width. Combining three translation components, three Euler-angle components, and one gripper command per arm yields a common 14-dimensional dual-arm interface. This standardization gives action labels consistent local-frame semantics across robot embodiments.

We check the standardized states and actions for non-finite values, consistent episode lengths, contiguous frame indices, and timestamps compatible with 10\,Hz. We also verify the temporal coverage of the synchronized head and bilateral wrist-camera streams. For continuous action dimensions, we estimate the 0.1\% and 99.9\% quantiles from sampled valid pre-training data and use these bounds to linearly normalize values to $[-1,1]$, clipping values outside the bounds. Binary gripper commands use fixed bounds of $[0,1]$ before the same affine normalization. Training windows remain within episode boundaries. When a future action chunk extends beyond an episode, the final action is repeated for padding, and padded positions are excluded from the action loss.
We pre-train LaST$_{1.0}$ for one epoch over the entire combined corpus, jointly optimizing action prediction and future-latent prediction with the same joint flow-matching loss as downstream fine-tuning (Eq.~\ref{eq:total_loss}), including the latent loss weight $\lambda_{\mathrm{lat}}=0.1$. All policy modules are updated end-to-end, while the VGGT-$\Omega$ target encoder remains frozen.

\subsubsection{Fine-tuning}
\label{sec:fine-tune}

We normalize all action dimensions to $[-1,1]$ using dataset quantile bounds, while applying z-normalization to the target latents. Both the action and latent branches are trained jointly via linear conditional flow matching \cite{lipman2022flow}. Let $u \sim \operatorname{Beta}(1.5, 1.0)$ denote the flow timestep, and $\boldsymbol{\epsilon} \sim \mathcal{N} (\mathbf{0}, \mathbf{I})$ be standard Gaussian noise. For a given generative target $\mathbf{Y} \in \{\mathbf{A}, \mathbf{Z}\}$, we construct the noised input $\mathbf{Y}_u$ and the target velocity $\mathbf{V}_Y$ as:
$\mathbf{Y}_u = (1-u)\mathbf{Y} + u\boldsymbol{\epsilon}, \mathbf{V}_Y = \boldsymbol{\epsilon} - \mathbf{Y},$
where $\mathbf{A}$ represents the action chunk $\mathbf{a}_{t+1:t+H}$ and $\mathbf{Z}$ denotes the latent targets. The backbone then predicts the corresponding flow velocities $\widehat{\mathbf{V}}_A$ and $\widehat{\mathbf{V}}_Z$. For a mini-batch of size $B$, we compute the action and latent Mean Squared Errors (MSEs) separately and optimize their weighted sum:
\begin{equation}
    \begin{aligned}
        \mathcal{L}_{\mathrm{act}} &=
        \frac{\sum_{b=1}^{B}\sum_{h=1}^{H} m_{b,h}
        \left\|\widehat{\mathbf{V}}_{A,b,h}-\mathbf{V}_{A,b,h}\right\|_2^2}
        {\max\!\left(1,\,d_A\sum_{b=1}^{B}\sum_{h=1}^{H}m_{b,h}\right)}, \quad
        \mathcal{L}_{\mathrm{lat}} = \frac{1}{B N_Z d_Z}\sum_{b=1}^{B}
        \left\|\widehat{\mathbf{V}}_{Z,b}-\mathbf{V}_{Z,b}\right\|_F^2, \quad
        \mathcal{L} = \mathcal{L}_{\mathrm{act}} + \lambda_{\mathrm{lat}}\mathcal{L}_{\mathrm{lat}}.
    \end{aligned}
    \label{eq:total_loss}
\end{equation}
Here, $d_A$ is the action dimension, and $N_Z$ and $d_Z$ are the number and dimension of target latent tokens. The action-loss mask $m_{b,h}\in\{0,1\}$ selects valid positions; when all positions participate in training, $m_{b,h}=1$. Thus, each branch averages squared errors over its contributing scalar elements before weighting, with $\lambda_{\mathrm{lat}}=0.1$.

\subsubsection{Inference}
\label{sec:inference}
At inference, following the decoupled topology, the latent pathway is omitted. Starting from pure Gaussian noise $\mathbf{A}_{u=1} \sim \mathcal{N}(\mathbf{0}, \mathbf{I})$, we synthesize the action chunk by querying the action vector field $\widehat{\mathbf{V}}_A$ and solving the flow ODE toward $u=0$ using standard Euler integration with $N_{\mathrm{flow}}=4$ steps.

\section{Experiments}

\subsection{Simulation Experiment}

\subsubsection{Benchmarks}
\label{sec:benchmarks}
We evaluate LaST$_{1.0}$ on the RLBench and LIBERO benchmarks to comprehensively validate its manipulation capabilities. For \textbf{RLBench} \cite{james2020rlbench}, we select 10 diverse tasks in CoppeliaSim utilizing a Franka Panda arm. Following standard protocols \cite{liu2025hybridvla,chen2025fast}, we use 100 expert trajectories per task for training. For \textbf{LIBERO} \cite{liu2023libero}, we utilize the official training datasets provided for its four suites (\emph{Spatial}, \emph{Object}, \emph{Goal}, and \emph{Long}). Each suite comprises 10 distinct tasks with 50 official expert demonstrations per task, totaling 500 trajectories per suite. Additionally, to assess generalization, we also evaluate on the \textbf{LIBERO-Plus} \cite{fei2025libero}. For this evaluation, we directly apply the weights trained on the LIBERO datasets without any further fine-tuning.

\subsubsection{Training and Evaluation Protocol}
\label{sec:evaluation_protocol}
We benchmark LaST$_{1.0}$ against recent state-of-the-art Vision-Language-Action (VLA) models, including OpenVLA-OFT \cite{kim2025fine}, $\pi_{0.5}$ \cite{intelligence2025pi05visionlanguageactionmodelopenworld}, HybridVLA \cite{liu2025hybridvla}, StarVLA-$\alpha$ \cite{ye2026starvla}, VLA-JEPA \cite{sun2026vla}, and LaST$_{0}$ \cite{liu2026last}, as well as representative World-Action Models (WAMs), including Fast-WAM \cite{yuan2026fast}, Cosmos Policy \cite{kim2026cosmos}, and LingBot-VA \cite{li2026causal}.
LaST$_{1.0}$ is trained using AdamW with a learning rate of $1 \times 10^{-4}$ on 8 NVIDIA RTX PRO 5000 GPUs. We train the model for 40k steps on RLBench. For LIBERO, we jointly train a single policy on the combined datasets of all four suites for 80k steps. All input images are resized to $256 \times 256$. On RLBench, the policy receives only the current front-view image as visual input; future wrist-view images are used exclusively to construct auxiliary latent supervision targets during training and are not required at inference. On LIBERO, the policy receives both current agent-view and wrist-view images.
During evaluation, we test the final checkpoint of each method. On RLBench, we perform 50 independent rollouts per task and report the success rate. On both LIBERO and LIBERO-Plus, we evaluate LaST$_{1.0}$ across three independent test runs, utilizing 50 rollouts per task for LIBERO and following the standard category-specific evaluation protocol for LIBERO-Plus.

\subsubsection{Results}
\label{sec:simulation_results}
As shown in Tables~\ref{tab:rlbench}, \ref{tab:libero}, and \ref{tab:libero_plus}, LaST$_{1.0}$ establishes a new state-of-the-art across all evaluated benchmarks. On RLBench, it achieves the highest average success rate of $92.0\%$, outperforming the strongest baseline, LaST$_0$ ($81.8\%$), by $10.2$ percentage points and eclipsing WAMs like Cosmos Policy ($76.4\%$). By scoring near-perfect success rates ($\ge 96.0\%$) on seven out of ten tasks, it shows that our encoder-free, geometry-aware reasoning preserves fine-grained spatial precision—an area where semantic-heavy VLAs (e.g., OpenVLA-OFT at $54.6\%$) inherently struggle. On the LIBERO benchmark, LaST$_{1.0}$ yields a leading $99.0\%$ average, notably maintaining $98.5\%$ on the challenging \textit{Long} suite, which demonstrates that our branch-decoupled regularization effectively injects geometric priors within the policy representation. Finally, under zero-shot transfer to LIBERO-Plus, LaST$_{1.0}$ delivers peak out-of-distribution robustness with the highest average of $86.3\%$. It excels across severe shifts like \textit{Sensor Noise} ($97.3\%$) and \textit{Language} ($89.4\%$), whereas baselines reliant on space-discretizing VAEs (Fast-WAM at $50.0\%$) suffer severe degradation. Crucially, LaST$_{1.0}$ significantly surpasses LaST$_0$ ($73.2\%$) by a striking margin of $13.1$ percentage points on LIBERO-Plus. This substantial improvement highlights that our proposed interaction-centric, geometry-aware latent world modeling provides far more robust and actionable cues for control than previous designs. These results demonstrate that learning interaction-aware geometric dynamics yields superior generalization over observation-complete video prediction.

\begin{table}[t]
\centering
\caption{\textbf{Results on RLBench.} All methods are trained in the multi-task setting, and we report average success rates (\%).}
\label{tab:rlbench}
\small
\setlength{\tabcolsep}{4.5pt}
\begin{tabular}{l*{11}{c}}
\toprule
\multirow{2}{*}{Method} & Close & Close & Toilet & Sweep to & Close & Phone & Take & Take Frame & Place Wine & Water & \multirow{2}{*}{Average} \\
& Box & Laptop Lid & Seat Down & Dustpan & Fridge & on Base & Umbrella & off Hanger & at Rack & Plants & \\
\midrule
OpenVLA-OFT~\cite{kim2025fine} & 96.0 & 70.0 & 80.0 & 50.0 & 84.0 & 38.0 & 44.0 & 20.0 & 38.0 & 26.0 & 54.6 \\
$\pi_{0.5}$~\cite{intelligence2025pi05visionlanguageactionmodelopenworld} & 90.0 & 96.0 & 94.0 & 84.0 & \textbf{100.0} & 18.0 & 28.0 & \textbf{86.0} & 76.0 & 38.0 & 71.0 \\
HybridVLA~\cite{liu2025hybridvla} & 92.0 & 96.0 & \textbf{100.0} & 90.0 & 96.0 & 44.0 & 58.0 & 68.0 & 48.0 & 44.0 & 73.6 \\
Fast-WAM~\cite{yuan2026fast} & 94.0 & 90.0 & 96.0 & 74.0 & \textbf{100.0} & 30.0 & 66.0 & 56.0 & 8.0 & 14.0 & 62.8 \\
Cosmos Policy~\cite{kim2026cosmos} & 94.0 & 94.0 & \textbf{100.0} & 60.0 & 76.0 & 78.0 & \textbf{80.0} & 70.0 & 54.0 & 58.0 & 76.4 \\
LaST$_{0}$~\cite{liu2026last} & 96.0 & 92.0 & \textbf{100.0} & 82.0 & 86.0 & 74.0 & 78.0 & 66.0 & 80.0 & 64.0 & 81.8 \\
\midrule
\rowcolor{LastTableHighlight}
\textbf{LaST$_{1.0}$ (Ours)} & \textbf{100.0} & \textbf{100.0} & \textbf{100.0} & \textbf{96.0} & \textbf{100.0} & \textbf{86.0} & 78.0 & 66.0 & \textbf{98.0} & \textbf{96.0} & \textbf{92.0} \\
\bottomrule
\end{tabular}
\end{table}

\begin{table}[tbp]
\centering
\caption{\textbf{Results on LIBERO.} We perform three test runs and report the average success rate (\%).}
\label{tab:libero}
\small
\setlength{\tabcolsep}{10pt}
\begin{tabular}{lccccc}
\toprule
Models & Spatial & Object & Goal & Long & Average \\
\midrule
OpenVLA-OFT~\cite{kim2025fine} & 97.6 & 98.4 & 97.9 & 94.5 & 97.1 \\
$\pi_{0.5}$~\cite{intelligence2025pi05visionlanguageactionmodelopenworld} & 98.8 & 98.2 & 98.0 & 92.4 & 96.9 \\
StarVLA-$\alpha$ \cite{ye2026starvla} & 99.0 & 99.8 & 98.5 & 94.1 & 97.9 \\
Fast-WAM~\cite{yuan2026fast} & 98.2 & \textbf{100.0} & 97.0 & 95.2 & 97.6 \\
Cosmos Policy~\cite{kim2026cosmos} & 98.1 & \textbf{100.0} & 98.2 & 97.6 & 98.5 \\
LingBot-VA \cite{li2026causal} & 98.5 & 99.6 & 97.2 & \textbf{98.5} & 98.5 \\
VLA-JEPA \cite{sun2026vla} & 96.2 & 99.6 & 97.2 & 95.8 & 97.2 \\
LaST$_{0}$~\cite{liu2026last} & \textbf{99.2} & 99.6 & 98.0 & 95.6 & 98.1 \\
\midrule
\rowcolor{LastTableHighlight}
\textbf{LaST$_{1.0}$ (Ours)} & 98.5 & 99.8 & \textbf{99.1} & \textbf{98.5} & \textbf{99.0} \\
\bottomrule
\end{tabular}
\end{table}

\subsection{Ablation Study}
\label{sec:ablation}

We conduct ablations on the same ten-task RLBench protocol, changing one factor at a time while keeping all other settings fixed.

\subsubsection{Latent World Modeling Design}
\noindent\textbf{(1) Choice of latent targets.}
We compare different future latent targets under a controlled policy interface, with target encoders frozen. VGGT-$\Omega$ uses its native $16$ registers, while alternative representations are pooled to $16$ tokens per future timestep and projected to the same backbone dimension. All other policy and optimization settings remain fixed.
As shown in Fig.~\ref{fig:ablation}(a), VGGT-$\Omega$ \cite{wang2026vggt} register targets achieve $92.0\%$ success, outperforming pooled full VGGT-$\Omega$ features ($89.6\%$), Uni3D-g \cite{zhou2024uni3d} point-cloud latents ($87.8\%$), Wan 2.2 VAE \cite{wan2025wan} video latents ($85.4\%$), DINOv3-H+ \cite{simeoni2025dinov3} image latents ($83.6\%$), and no latent supervision ($81.8\%$). Compared with the Uni3D-g target variant inspired by LaST$_0$ \cite{liu2026last}, our method improves success by $4.2$ percentage points without explicit point-cloud inputs.

\noindent\textbf{(2) Geometric content of predicted latents.}
To assess whether the predicted future latents encode control-relevant geometry, we perform timestep-aligned linear probing. For probing, we retain the latent branch and generate future latents conditioned only on the current observations and language instruction. During robot deployment, this branch can be omitted for action-only generation. We freeze the policy and fit a linear ridge probe on the policy-inferred future latents from the training demonstrations. For evaluation, we collect 50 additional demonstrations per task as a held-out test set and report the probing error on their inferred future latents. The probe predicts the relative position between the end-effector and a contact-relevant object point at each predicted future timestep,
$r_{t+\delta_f} = p^{\mathrm{obj}}_{t+\delta_f} - p^{\mathrm{ee}}_{t+\delta_f}, \ \delta_f=fH/F,\ f\in\{1,\ldots,F\}$,
where $F$ denotes the number of sampled future timesteps and $H$ denotes the action horizon. For each task, the ground-truth contact-relevant object point is determined in the simulator based on the waypoints of the corresponding execution trajectory. Each predicted latent is probed against the geometry at its matched timestep in a common coordinate frame. We report the three-dimensional Euclidean error averaged over the predicted future timesteps.
This metric directly captures the robot--object interaction geometry relevant to manipulation.
As shown in Table~\ref{tab:future_geometry_probe}, VGGT-$\Omega$ registers achieve the lowest error ($0.23$ cm), followed by pooled VGGT-$\Omega$ full features ($0.29$ cm), Uni3D-g ($0.77$ cm), Wan 2.2 VAE ($1.88$ cm), and DINOv3-H+ ($2.69$ cm). The error ordering agrees with manipulation success, suggesting that future latent target choice affects the accessibility of manipulation-relevant spatial relations.

\begin{table}[tbp]
\centering
\caption{\textbf{Results on LIBERO-Plus.} We perform three test runs and report the average success rate (\%).}
\label{tab:libero_plus}
\small
\setlength{\tabcolsep}{8pt}
\begin{tabular}{lcccccccc}
\toprule
Models & Camera & Robot & Language & Light & Background & Noise & Layout & Average \\
\midrule
OpenVLA-OFT~\cite{kim2025fine} & 56.4 & 31.9 & 79.5 & 88.7 & 93.3 & 75.8 & 74.2 & 69.6 \\
$\pi_{0.5}$~\cite{intelligence2025pi05visionlanguageactionmodelopenworld} & 78.4 & 73.6 & 80.8 & 96.2 & 94.1 & 89.0 & 84.5 & 84.4 \\
StarVLA-$\alpha$ \cite{ye2026starvla} & 48.7 & 63.4 & 86.8 & 95.8 & \textbf{94.6} & 75.0 & 80.2 & 76.1 \\
Fast-WAM~\cite{yuan2026fast} & 16.4 & 44.5 & 68.9 & 78.2 & 53.7 & 37.7 & 60.7 & 50.0 \\
Cosmos Policy~\cite{kim2026cosmos} & 75.8 & 63.3 & 81.7 & \textbf{96.5} & 88.9 & 92.7 & 82.2 & 82.2 \\
LingBot-VA \cite{li2026causal} & 40.9 & \textbf{83.0} & 86.4 & 82.3 & 53.1 & 64.4 & 76.2 & 69.5 \\
VLA-JEPA \cite{sun2026vla} & 63.3 & 67.1 & 85.4 & 95.6 & 93.6 & 66.3 & 85.1 & 78.0 \\
LaST$_{0}$~\cite{liu2026last} & \textbf{81.4} & 36.5 & 80.2 & 86.4 & 85.7 & 79.8 & 68.9 & 73.2 \\
\midrule
\rowcolor{LastTableHighlight}
\textbf{LaST$_{1.0}$ (Ours)} & 77.9 & 68.9 & \textbf{89.4} & 94.7 & 93.1 & \textbf{97.3} & \textbf{87.3} & \textbf{86.3} \\
\bottomrule
\end{tabular}
\end{table}

\noindent\textbf{(3) Observation views of latent targets.}
Figure~\ref{fig:ablation}(b) shows that wrist-only targets
($92.0\%$) outperform external-view ($87.6\%$) and both-view
($89.4\%$) targets.
Together with the probing results, this supports concentrating
future supervision on the robot--object interaction region.
Adding an external view does not improve performance in this setting,
suggesting that broader scene coverage is not necessarily a more
effective auxiliary prediction target for manipulation.

\noindent\textbf{(4) Predictive supervision and deployment.}
We also distinguish future prediction from current-state
representation supervision.
Under the single-timestep setting, predicting a future geometry-aware
latent achieves $88.2\%$ success ($F=1$ in
Table~\ref{tab:ablation_compact}), compared with $82.4\%$ when
the target is extracted from the current observation.
This result supports the value of predictive supervision beyond
aligning the policy with current geometric features.
Finally, the decoupled mask yields a $92.0\%$ success rate, compared to $89.2\%$ for coupled attention (Fig.~\ref{fig:ablation}(c)). In the coupled variant, action tokens can attend to latent tokens during both training and inference, and the latent branch is retained at deployment. The decoupled design makes action generation independent of future-latent tokens, allowing the latent branch to be omitted at inference while retaining the benefits of predictive supervision during training.

\subsubsection{Architecture Design}
\begin{figure}[tbp]
    \centering
    \includegraphics[width=\textwidth]{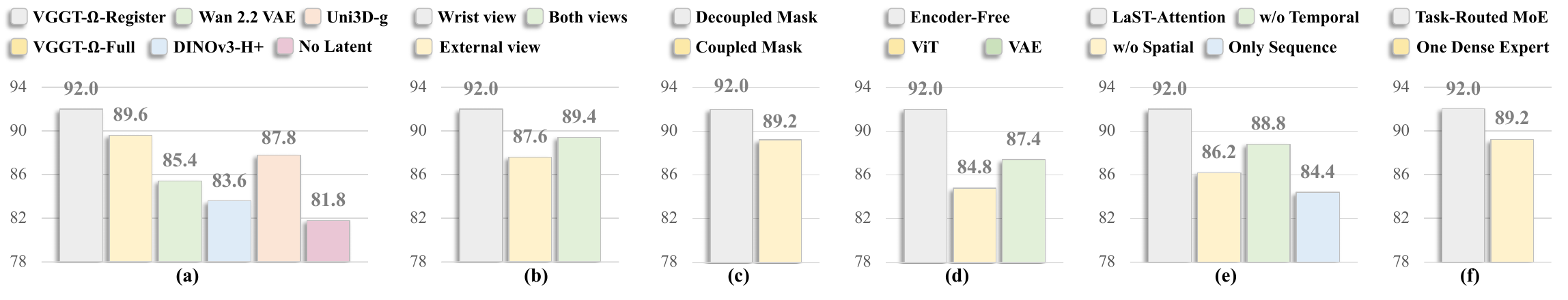}
    \caption{\textbf{Ablation studies of LaST$_{1.0}$ on RLBench.} \textbf{(a--c)} Ablations on the latent world modeling design, including future latent target, observation views, and the branch-decoupled mask. \textbf{(d--f)} Ablations on the architecture design, comparing the encoder-free interface, LaST-Attention with 4D-RoPE, and the Task-Routed MoE. We report the average success rate (\%).}
    \label{fig:ablation}
\end{figure}

\begin{figure}[t]
    \centering
    \includegraphics[width=0.8\linewidth]{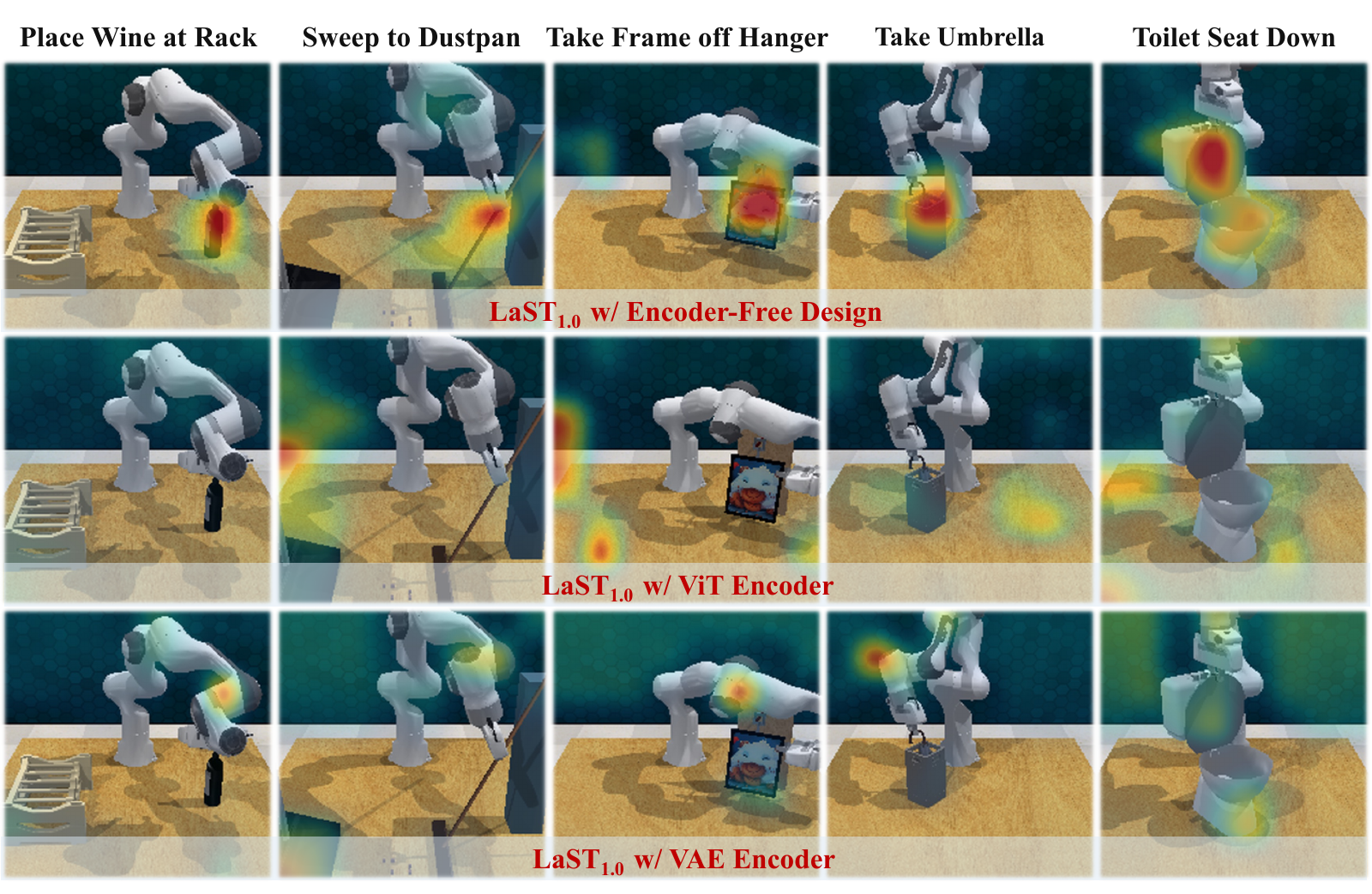}
    \caption{\textbf{Action-to-Vision Attention Maps.} Qualitative examples for the encoder-free, ViT, and VAE variants. In these examples, the encoder-free variant shows localized attention around task-relevant objects and interaction regions.}
    \label{fig:attention}
\end{figure}

Regarding the visual input representation, our encoder-free interface achieves $92.0\%$, outperforming a SigLIP2 \cite{tschannen2025siglip} vision encoder ($84.8\%$) and a reconstruction-oriented Wan2.2 VAE \cite{wan2025wan} ($87.4\%$) as shown in Fig.~\ref{fig:ablation}(d). The qualitative attention visualizations in Fig.~\ref{fig:attention} show localized action-to-vision attention around manipulated objects in our encoder-free variant, suggesting a stronger focus on control-relevant regions.
We next evaluate LaST-Attention with 4D-RoPE in Fig.~\ref{fig:ablation}(e). The full 4D formulation achieves $92.0\%$. Removing the spatial coordinates $(y,x)$ while retaining sequence and time $(n,\tau)$ reduces success to $86.2\%$, whereas removing the temporal coordinate $\tau$ while retaining sequence and space $(n,y,x)$ yields $88.8\%$. Retaining only the sequence coordinate $n$ further reduces success to $84.4\%$.
Finally, the Task-Routed MoE ($92.0\%$) outperforms a single dense FFN expert ($89.2\%$; Fig.~\ref{fig:ablation}(f)), whose parameter count approximately matches that of the three routed FFN experts combined. This result supports the benefit of role-specific computation under a comparable total parameter budget.

\subsubsection{Hyperparameters Design}
Table~\ref{tab:ablation_compact} shows that a moderate latent weight performs best. $\lambda_{\mathrm{lat}}=0.1$ achieves $92.0\%$, compared with $81.8\%$, $89.2\%$, and $87.8\%$ for $0$, $0.5$, and $1.0$. With this weight fixed, using $F=1,2,4,$ and $8$ future frames gives $88.2\%$, $89.8\%$, $92.0\%$, and $90.4\%$, respectively. We therefore adopt $\lambda_{\mathrm{lat}}=0.1$ and $F=4$ for LaST$_{1.0}$ by default.

\begin{table}[tbp]
    \centering
    \small
    \caption{\textbf{Future latent probing results on RLBench.}}
    \label{tab:future_geometry_probe}
    \setlength{\tabcolsep}{8pt}
\begin{tabular}{lccccc}
        \toprule
        & VGGT-$\Omega$-Register & VGGT-$\Omega$-Full & Uni3D-g & Wan2.2 VAE & DINOv3 \\
        \midrule
        Error (cm) $\downarrow$
        & $\mathbf{0.23}$ & $0.29$ & $0.77$ & $1.88$ & $2.69$ \\
        \bottomrule
    \end{tabular}
\end{table}

\begin{table}[t]
    \centering
    \small
    \caption{\textbf{Hyperparameter Ablations on RLBench.}
    We ablate the latent loss weight $\lambda_{\mathrm{lat}}$ and the sampled frames $F$.}
    \label{tab:ablation_compact}
    \setlength{\tabcolsep}{8pt}
    \renewcommand{\arraystretch}{1.05}
        \begin{tabular}{lccccccccc}
            \toprule
            \multirow{2}{*}[-0.6ex]{Value} & \multicolumn{4}{c}{Latent Loss Weight $\lambda_{\mathrm{lat}}$}
            & &
            \multicolumn{4}{c}{Future Frames $F$} \\
            \cmidrule{2-5} \cmidrule{7-10}
                & 0 & 0.1 & 0.5 & 1.0
                & &
                1 & 2 & 4 & 8 \\
            \midrule
            Success rates (\%)
                & 81.8 & \textbf{92.0} & 89.2 & 87.8
                & &
                88.2 & 89.8 & \textbf{92.0} & 90.4 \\
            \bottomrule
        \end{tabular}
\end{table}

\subsection{Real-World Experiment}

\subsubsection{Data Collection and Task Definition}
\label{sec:data_collection}
To evaluate LaST$_{1.0}$ across diverse embodiments and task horizons, we design five real-world multi-stage manipulation tasks. Data collection strategies are tailored to the control complexity of each task.

\noindent\textbf{(1) Dual-arm Franka with Grippers:} We utilize a SpaceMouse to collect 100 expert trajectories per task. The evaluation includes two tasks: \textit{Put Object in Bag} (S1: place the object inside, and S2: zip the bag) and \textit{Put Object in Drawer} (S1: open the top drawer, S2: place the object inside, and S3: close the drawer).

\noindent\textbf{(2) Dual-arm Tianji with Grippers:} For fine-grained, long-horizon manipulation, we collect 300 demonstrations per task via a PICO VR teleoperation platform. The tasks are \textit{Parcel-Box Packing} (S1: place the item into the box, S2: fold the four flaps, and S3: seal the box with tape) and \textit{Assemble Truck Toy} (S1: position the wheel, S2: place the white accessory, S3: fasten with an electric drill, and S4: flip the truck upright).

\noindent\textbf{(3) Dual-arm Tianji with Sharpa Hands:} To validate dexterous hand control, we capture 100 trajectories using Manus motion-capture gloves for the \textit{Stack Cups} task (S1: stack the right cup onto the middle cup, followed by the left cup (S2)).

\subsubsection{Training and Evaluation Protocol}
\label{sec:evaluation_protocol_real}
We compare LaST$_{1.0}$ against three state-of-the-art baselines: two VLA models ($\pi_{0.5}$ \cite{intelligence2025pi05visionlanguageactionmodelopenworld} and LaST$_{0}$ \cite{liu2026last}) and one WAM (Cosmos Policy \cite{kim2026cosmos}). For a rigorous and fair comparison, all policies receive identical visual inputs consisting of three camera streams: two wrist views and one static third-person exterior view, all resized to $256 \times 256$ resolution. All baseline policies are trained on our exact real-world demonstration datasets using their officially released training scripts. LaST$_{1.0}$ is trained for $10\text{k}$ steps per task in real-world experiments, using the same setting as in the simulation.
During evaluation, each policy is tested on 20 real-world rollouts per task under varying initial object poses and spatial arrangements. We decompose every task into its constituent stages (as defined in Sec.~\ref{sec:data_collection}). A stage is counted as successful only if all preceding stages and the current stage are completed successfully. Its success rate is the fraction of all 20 rollouts that meet this criterion. The full-task success rate therefore equals the success rate of the final stage. We report the stage-level average across all 14 stages and the task-level average across the five full-task success rates.

\begin{table}[t]
\centering
\caption{\textbf{Results on Real-World Tasks.} We decompose each task into subtasks and record the success rates (\%). Inference time (ms) is measured per model inference on an NVIDIA RTX 4090 GPU under identical settings.}
\label{tab:real_world}
\small
\setlength{\tabcolsep}{4pt}
\begin{tabular}{l*{14}{c}@{\hspace{9pt}}ccc}
\toprule
\multirow{3}{*}{Method} & \multicolumn{5}{c}{Franka + Gripper} & \multicolumn{7}{c}{Tianji + Gripper} & \multicolumn{2}{c}{Tianji + Hand} & \multicolumn{3}{c}{Overall} \\
\cmidrule(lr){2-6}\cmidrule(lr){7-13}\cmidrule(lr){14-15}\cmidrule(l){16-18}
& \multicolumn{2}{c}{\shortstack{Put Object\\in Bag}} & \multicolumn{3}{c}{\shortstack{Put Object\\in Drawer}} & \multicolumn{3}{c}{\shortstack{Parcel-Box\\Packing}} & \multicolumn{4}{c}{\shortstack{Assemble\\Truck Toy}} & \multicolumn{2}{c}{\shortstack{Stack\\Cups}} & \multirow{2}{*}{\shortstack{Stage\\Average}} & \multirow{2}{*}{\shortstack{Task\\Average}} & \multirow{2}{*}{\shortstack{Inference\\Time (ms)}} \\
\cmidrule(lr){2-3}\cmidrule(lr){4-6}\cmidrule(lr){7-9}\cmidrule(lr){10-13}\cmidrule(lr){14-15}
& S1 & S2 & S1 & S2 & S3 & S1 & S2 & S3 & S1 & S2 & S3 & S4 & S1 & S2 & & & \\
\midrule
$\pi_{0.5}$ \cite{intelligence2025pi05visionlanguageactionmodelopenworld} & 75.0 & 50.0 & \textbf{90.0} & 70.0 & 65.0 & \textbf{90.0} & 70.0 & 20.0 & \textbf{85.0} & 10.0 & 5.0 & 5.0 & 60.0 & 40.0 & 52.5 & 36.0 & 60.4 \\
Cosmos Policy \cite{kim2026cosmos} & 80.0 & 55.0 & 85.0 & 45.0 & 35.0 & 70.0 & 60.0 & 10.0 & 70.0 & 5.0 & 0.0 & 0.0 & 60.0 & \textbf{50.0} & 44.6 & 30.0 & 229.0 \\
LaST$_{0}$ \cite{liu2026last} & 80.0 & 60.0 & \textbf{90.0} & 60.0 & 60.0 & 75.0 & 70.0 & 15.0 & 75.0 & 20.0 & 15.0 & 5.0 & 65.0 & 35.0 & 51.8 & 35.0 & 138.3 \\
\midrule
\rowcolor{LastTableHighlight}
\textbf{LaST$_{1.0}$ (Ours)} & \textbf{95.0} & \textbf{70.0} & \textbf{90.0} & \textbf{80.0} & \textbf{75.0} & \textbf{90.0} & \textbf{75.0} & \textbf{45.0} & \textbf{85.0} & \textbf{45.0} & \textbf{35.0} & \textbf{30.0} & \textbf{75.0} & \textbf{50.0} & \textbf{67.1} & \textbf{54.0} & \textbf{54.7} \\
\bottomrule
\end{tabular}
\end{table}

\begin{figure}[tbp]
    \centering
    \includegraphics[width=\textwidth]{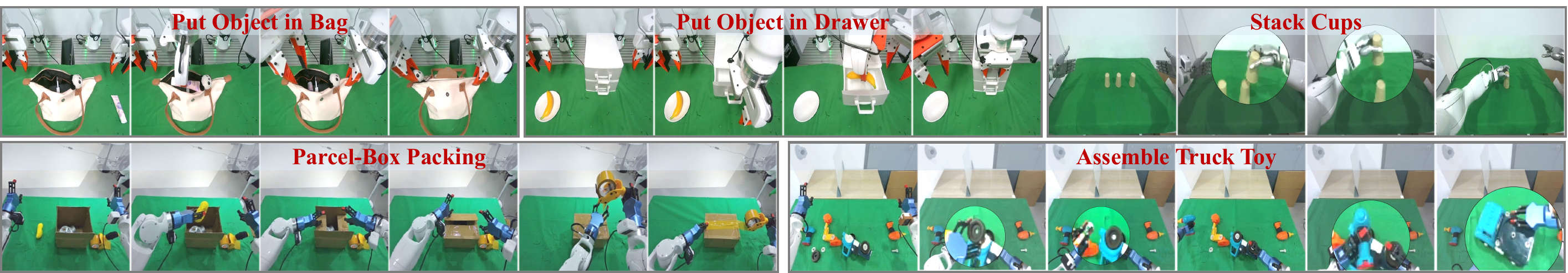}
    \caption{\textbf{Visualization of LaST$_{1.0}$ performing diverse tasks across different robotic embodiments in real-world executions.}}
    \label{fig:real_world}
\end{figure}

\subsubsection{Quantitative and Qualitative Results}
\label{sec:quan_qual_res}
As shown in Table~\ref{tab:real_world} and visualized in Fig.~\ref{fig:real_world}, LaST$_{1.0}$ achieves the highest or joint-highest task success rate among the compared methods on all five real-world tasks. It achieves stage-level and task-level average success rates of $67.1\%$ and $54.0\%$, respectively, outperforming $\pi_{0.5}$ ($52.5\%$/$36.0\%$), LaST$_{0}$ ($51.8\%$/$35.0\%$), and Cosmos Policy ($44.6\%$/$30.0\%$). The higher full-task success rates indicate stronger performance on multi-stage manipulation. In the four-stage \textit{Assemble Truck Toy} task, LaST$_{1.0}$ reaches a second-stage success rate of $45.0\%$, compared with at most $20.0\%$ for the baselines. It also completes the full task in $30.0\%$ of rollouts, compared with at most $5.0\%$ for the baselines. On the dexterous \textit{Stack Cups} task using Sharpa hands, LaST$_{1.0}$ matches the strongest baseline with a full-task success rate of $50.0\%$.
Beyond execution success, LaST$_{1.0}$ requires $54.7$~ms per model inference. This inference time is lower than those of LaST$_0$ ($138.3$~ms) and Cosmos Policy ($229.0$~ms), demonstrating efficient action generation with the encoder-free interface and the latent branch omitted at inference.

\begin{table}[t]
\centering
    \small
\caption{\textbf{Generalization Analysis.} Task-level success rates (\%) for completing both stages of ``Put Object in Bag'', evaluated over 20 rollouts per model per setting. Parenthesized values indicate relative decreases from the original-setting success rate.}
\label{tab:gene}
\setlength{\tabcolsep}{8pt}
\begin{tabular}{lcccc}
\toprule
Model & Original & Object & Background & Position \\
\midrule
$\pi_{0.5}$ \cite{intelligence2025pi05visionlanguageactionmodelopenworld} & 50.0 & 35.0 {\color{gray}($\downarrow$ 30\%)} & 50.0 {\color{gray}($\downarrow$ 0\%)} & 20.0 {\color{gray}($\downarrow$ 60\%)} \\
Cosmos Policy \cite{kim2026cosmos} & 55.0 & 20.0 {\color{gray}($\downarrow$ 64\%)} & 35.0 {\color{gray}($\downarrow$ 36\%)} & 10.0 {\color{gray}($\downarrow$ 82\%)} \\
LaST$_{0}$ \cite{liu2026last} & 60.0 & 45.0 {\color{gray}($\downarrow$ 25\%)} & 45.0 {\color{gray}($\downarrow$ 25\%)} & 35.0 {\color{gray}($\downarrow$ 42\%)} \\
\rowcolor{LastTableHighlight}
\textbf{LaST$_{1.0}$ (Ours)} & \textbf{70.0} & \textbf{55.0} {\color{gray}($\downarrow$ 21\%)} & \textbf{70.0} {\color{gray}($\downarrow$ 0\%)} & \textbf{50.0} {\color{gray}($\downarrow$ 29\%)} \\
\bottomrule
\end{tabular}
\end{table}

\subsubsection{Generalization Analysis}
\label{sec:generalization}

To evaluate real-world robustness, we design three out-of-distribution (OOD) scenarios for the dual-arm ``Put Object in Bag'' task (Fig.~\ref{fig:teaser}(d)). Specifically, we introduce \textbf{(1) Object Generalization} by replacing the training object (facial cleanser) with a novel item of different shape and texture (banana); \textbf{(2) Background Generalization} by scattering multiple distractor objects across the workspace to create visual interference; and \textbf{(3) Position Generalization} by placing the target object in a spatial position outside the training distribution. For each model, we directly use its checkpoint from the original task without additional training or fine-tuning and perform 20 rollouts in each OOD setting, matching the original-setting evaluation count.
As shown in Table~\ref{tab:gene}, LaST$_{1.0}$ achieves the highest original-setting success rate (70.0\%) and the highest success rate in each evaluated OOD setting.
Relative to the original setting, LaST$_{1.0}$ shows a 21\% decrease under object generalization and no decrease (0\%) under background generalization, compared with relative decreases of 64\% and 36\% for Cosmos Policy, respectively.
Under position generalization, LaST$_{1.0}$ retains a 50.0\% success rate (a 29\% relative decrease), while the baselines show relative decreases of over 40\%. Overall, across these three settings, LaST$_{1.0}$ achieves a 58.3\% average success rate, outperforming LaST$_0$ (41.7\%) and Cosmos Policy (21.7\%). These results demonstrate higher success rates across the tested object, background, and position shifts.

\section{Conclusion}
\label{sec:conclusion}

In this paper, we presented LaST$_{1.0}$, a unified latent World-Action Model that reformulates predictive world modeling for manipulation as control-centric future prediction in latent space. Rather than modeling an observation-complete video future, LaST$_{1.0}$ predicts compact, geometry-aware latents focused on future robot--object interactions, providing predictive supervision relevant for control while maintaining action-only deployment. To support this paradigm, we redesigned the visual input interface and attention mechanism with a vision-encoder-free architecture and LaST-Attention for structured spatiotemporal interactions across modalities.
Extensive evaluations across simulation and real-world settings demonstrate state-of-the-art performance and strong generalization, validating control-centric latent world modeling as an effective approach for robust manipulation.

\bibliographystyle{unsrtnat}
\bibliography{reference}
\end{document}